\documentclass{article}
\usepackage{arxiv}

\usepackage[utf8]{inputenc} 
\usepackage[T1]{fontenc}    
\usepackage{hyperref}       
\usepackage{url}            
\usepackage{booktabs}       
\usepackage{amsfonts}       
\usepackage{nicefrac}       
\usepackage{microtype}      
\usepackage{lipsum}
\usepackage{graphicx}
\usepackage{enumitem}
\usepackage{diagbox}
\graphicspath{ {./images/} }
\usepackage{siunitx}
\usepackage{threeparttable}

\usepackage[round]{natbib} 
\usepackage{xcolor}
\usepackage{caption}
\usepackage{hyperref}
\usepackage{cleveref}
\crefname{figure}{Fig.}{Figs.}
\hypersetup{
    colorlinks=true,
    linkcolor=blue,    
    citecolor=blue,    
    urlcolor=blue      
}

\newcommand{\RN}[1]{\uppercase\expandafter{\romannumeral#1}}

\title{A Lightweight Large Language Model-Based Multi-Agent System for 2D Frame Structural Analysis}

\author{
Ziheng Geng$^{1,*}$,
Jiachen Liu$^{2,*}$,
Ran Cao$^{3}$,
Lu Cheng$^{4}$,
Haifeng Wang$^{5}$,
Minghui Cheng$^{1,6\dagger}$\\
\\
$^{1}$Department of Civil and Architectural Engineering, University of Miami, Coral Gables, FL 33146, USA\\
$^{2}$Department of Electrical and Computer Engineering, University of Miami, Coral Gables, FL 33146, USA\\
$^{3}$College of Civil Engineering, Hunan University, Changsha, 410082, China\\
$^{4}$Department of Computer Science, University of Illinois Chicago, Chicago, IL 60607, USA\\
$^{5}$Department of Civil and Environmental Engineering, Washington State University, Pullman, WA, 99164, USA\\
$^{6}$School of Architecture, University of Miami, Coral Gables, FL 33146, USA\\
\\
$^{*}$Equal contribution.\\
$^{\dagger}$Corresponding author: \texttt{minghui.cheng@miami.edu}
}

\begin{document}
\maketitle
\begin{abstract}
Large language models (LLMs) have recently been used to empower autonomous agents in engineering, significantly improving automation and efficiency in labor-intensive workflows. However, their potential remains underexplored in structural engineering, particularly for finite element modeling tasks requiring geometric modeling, complex reasoning, and domain knowledge. To bridge this gap, this paper develops a LLM-based multi-agent system to automate finite element modeling of 2D frames . The system decomposes structural analysis into subtasks, each managed by a specialized agent powered by the lightweight Llama-3.3 70B Instruct model. The workflow begins with a Problem Analysis Agent, which extracts geometry, boundary, and material parameters from the user’s input. Next, a Geometry Agent incrementally derives node coordinates and element connectivity by applying expert-defined rules. These structured outputs are converted into executable OpenSeesPy code by a Translation Agent and refined by a Model Validation Agent through consistency checks. Then, a Load Agent applies load conditions into the assembled structural model. Experimental evaluations on 20 benchmark problems demonstrate that the system achieves accuracy over 80\% in most cases across 10 repeated trials, outperforming Gemini-2.5 pro and ChatGPT-4o. 

\end{abstract}

\begin{quote}
\textbf{Keywords:} \textnormal{Large Language Model, Agent, Structural Analysis, Reliability, Finite Element Modeling}
\end{quote}

\section{Introduction}
Large language models (LLMs) have emerged as transformative tools for natural language processing, demonstrating unprecedented proficiency in comprehending, generating, and reasoning with human language. State-of-the-art LLMs, such as GPT-5 \citep{openai2025gpt5intro}, Gemini 2.5 \citep{comanici2025gemini}, and DeepSeek-R1 \citep{guo2025deepseek}, have exhibited exceptional capabilities across a vast spectrum of tasks, including content summarization \citep{ravaut2023context, zhang2024comprehensive}, mathematical reasoning \citep{ahn2024large, didolkar2024metacognitive}, and code generation \citep{jiang2024survey, liu2024exploring}, underscoring their versatility as general-purpose problem solvers. Despite these advances, a critical challenge lies in adapting such models to specialized domains that demand strict adherence to domain-specific rules, technical precision, and interpretability. This challenge has led to extensive research into techniques that enhance domain alignment. Prominent strategies include fine-tuning on curated and task-specific dataset \citep{jeong2024fine, parthasarathy2024ultimate}, in-context learning to derive reasoning patterns from domain examples \citep{zhang2024large, liang2025integrating}, and retrieval-augmented generation (RAG) to ground model outputs in external knowledge databases \citep{fan2024survey, li2024enhancing}. Collectively, these techniques adapt LLMs to specialized tasks that require domain knowledge and skills, paving the way for their integration into real-world scenarios.

In engineering and related applied domains, a growing research direction involves the development of LLMs into autonomous computational agents. Within such agents, the LLM functions as a central reasoning engine, capable of planning tasks, invoking external tools, and executing actions in interactive environments. By integrating modular toolkits, LLMs can translate natural language instructions into executable workflows. Recent studies have demonstrated this paradigm across diverse domains. For example, in geospatial science, \cite{akinboyewa2025gis} developed GIS Copilot, an agent that autonomously plans spatial analysis workflows and generates QGIS code by leveraging comprehensive documentation of GIS tools and parameters. In transportation engineering, \cite{li2024chatsumo} introduced ChatSUMO, which leverages LLMs to automate the generation and simulation of both abstract and real-world urban mobility scenarios. In fluid dynamics, \cite{pandey2025openfoamgpt} proposed OpenFOAMGPT, which employs RAG to embed domain knowledge to automate computational fluid dynamics simulations. Similarly, \cite{xu2024llm} presented FoamPilot for fire engineering, enabling automated simulations of fire dynamics and suppression through code insight, case configuration, and simulation execution. Together, these applications highlight the transformative potential of LLM-empowered agents to revolutionize workflows across scientific and industrial domains.

Engineering tasks, however, are inherently complex and typically require coordinated efforts among specialized experts to deliver complete solutions. Inspired by this collaborative workflow, LLM-based multi-agent systems have been recognized as a promising approach to manage complex problems in engineering practice \citep{han2024llm, li2024survey}. These systems adopt a divide-and-conquer strategy, decomposing large tasks into subtasks that are delegated to specialized agents. Recent studies illustrate their potential in domain-specific applications. For instance, \cite{elrefaie2025ai} developed a multi-agent framework comprising styling, computer-aided design (CAD), meshing, and simulation agents to automate car design considering both aesthetics and aerodynamic. \cite{chen2025multi} introduced a system with a task dispatcher, structural design agent, design evaluation agent, and expert consultation agent to achieve code-compliant design of reinforced concrete structures. \cite{xie2025rag} proposed a framework including a task orchestrator, user profile agent, planning agent, and analyst agent to support decision-making for natural hazard resilience and adaptation. These successful applications highlight the effectiveness of multi-agent systems in addressing engineering challenges. However, their potential remains largely unexplored in structural engineering due to the unique challenges of geometric reasoning, sequential modeling, and strict adherence to domain rules.

This study builds upon our previous work, where an LLM-empowered agent was developed to automate beam analysis in OpenSeesPy \citep{liu2025large}. A natural progression is to extend from 1D beams to 2D frame structures. However, this extension presents non-trivial challenges, as frame analysis requires spatial reasoning and topological consistency, which remain well-documented limitations of current LLMs \citep{tang2025lego, zhang2025call}. To address these challenges, an LLM-based multi-agent system is developed to automate the generation and execution of OpenSeesPy code for frame structural analysis. For open-source accessibility and deployment scalability, a lightweight LLM, Llama-3.3 70B Instruct model, is adopted as the core reasoning engine. Specifically, the system orchestrates a workflow comprising five specialized agents: a Problem Analysis Agent to extract geometry, material properties, and boundary conditions; a Geometry Agent to define nodes and elements in a stepwise manner; a Code Translation Agent to convert natural-language instructions into executable code; a Model Validation Agent to ensure consistency through duplication checks; and a Load Agent to apply load patterns. A benchmark dataset containing 20 frame problems is created to evaluate the performance of the proposed multi-agent system. The results demonstrate that the system accurately generates executable code and outperforms state-of-the-art LLMs, offering a pathway toward enhanced automation and efficiency in structural engineering practice.

\section{Benchmark}
\label{sec:headings}

\subsection{Dataset overview}
Frame structures is a fundamental load-bearing component in modern civil infrastructure, underpinning applications that range from large-scale industrial plants to high-rise buildings. Their design depends on accurate modeling of structural behavior, including axial, shear, and bending responses. In engineering practice, frame structural analysis is commonly performed using finite element modeling (FEM) tools, such as the open-source platforms OpenSees \citep{mckenna2011opensees} and commercial software packages including Ansys \citep{ansys}, Abaqus \citep{abaqus}, and SAP2000 \citep{sap2000}. While these software environments provide powerful modeling and simulation capabilities, their effective use requires domain knowledge in structural engineering and proficiency in navigating complex software interfaces. Additionally, the modeling process involves defining geometry, boundary conditions, material properties, and load patterns, all of which demand extensive manual effort that is time-consuming and error-prone. These limitations underscore the urgent need to leverage LLMs for automating frame structural analysis. 

\begin{figure*}[htbp]
\centering
\includegraphics[width=0.89\textwidth]{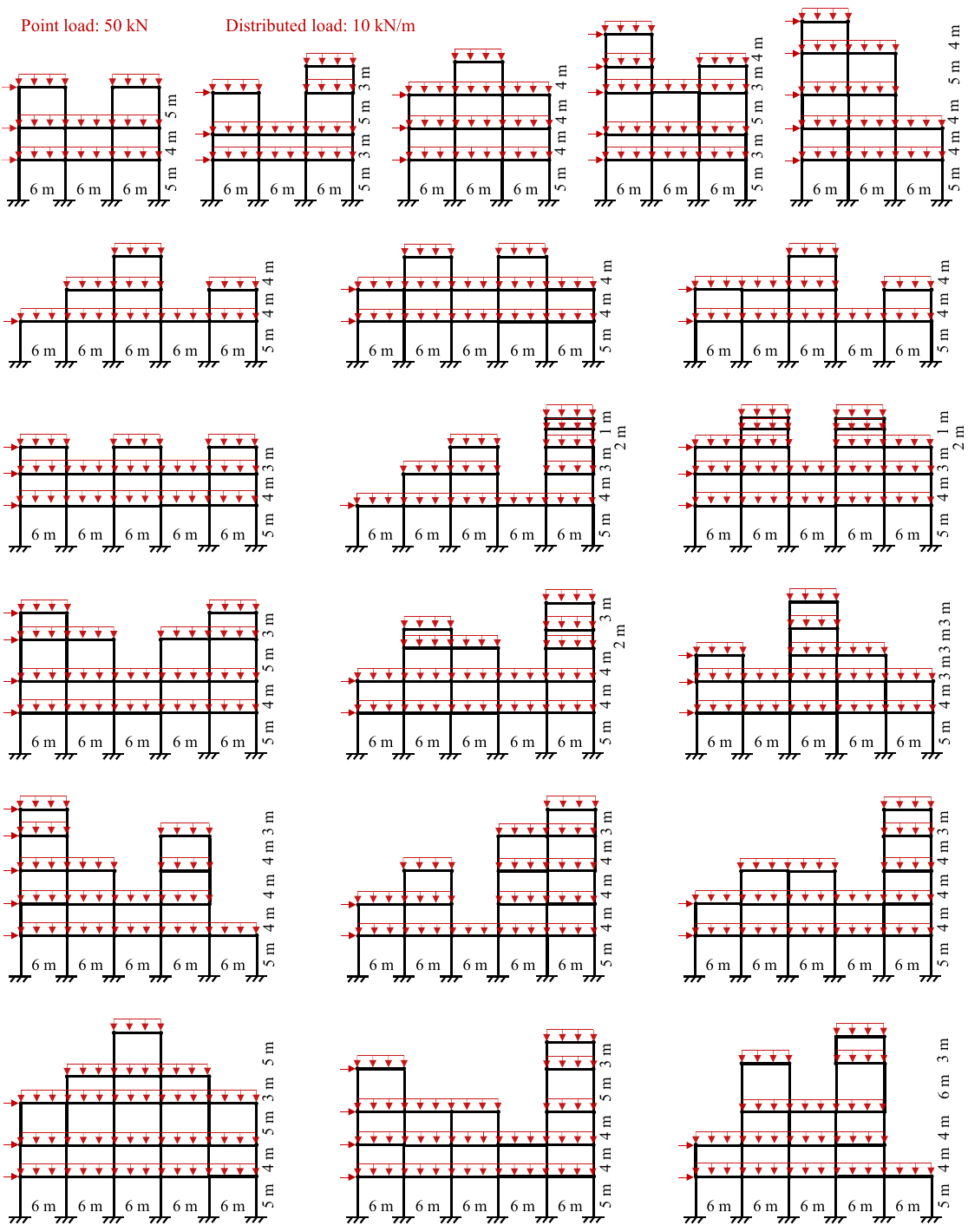}
\caption{A benchmark dataset comprising twenty representative frame structural analysis problems.}
\label{Figure1}
\end{figure*}

To evaluate the capabilities of LLMs in frame structural analysis, a benchmark dataset including 20 distinct frame problems is developed, as shown in \cref{Figure1}. The dataset integrates structural topologies from real-world cases with synthetically generated examples to ensure representativeness and generalizability. For instance, the 3-2-3 frame (three stories in the first bay, two in the second, and three in the third) is derived from the Notre Dame de Paris, while the 3-4-3 frame is adapted from the Mollien Pavilion of the Louvre Museum. To complement these realistic cases, additional frames are generated randomly. In particular, for five-bay frames, the number of stories in each bay is sampled independently from one to five, thereby yielding diverse structural configurations. For all frames in the dataset, consistent boundary and load conditions are applied. Specifically, fixed supports are assigned along the base, while the loads consist of two components: a point load of 50 kN applied rightward at the top of each story in the leftmost bay, and a uniformly distributed load of 10 kN/m applied downward along all beams. Collectively, these problems enable a holistic evaluation of LLM performance across the entire structural analysis workflow. 

\subsection{Performance evaluation of Llama-3.3 70B Instruct model}
\begin{figure*}[htbp]
\centering
\includegraphics[width=0.9\textwidth]{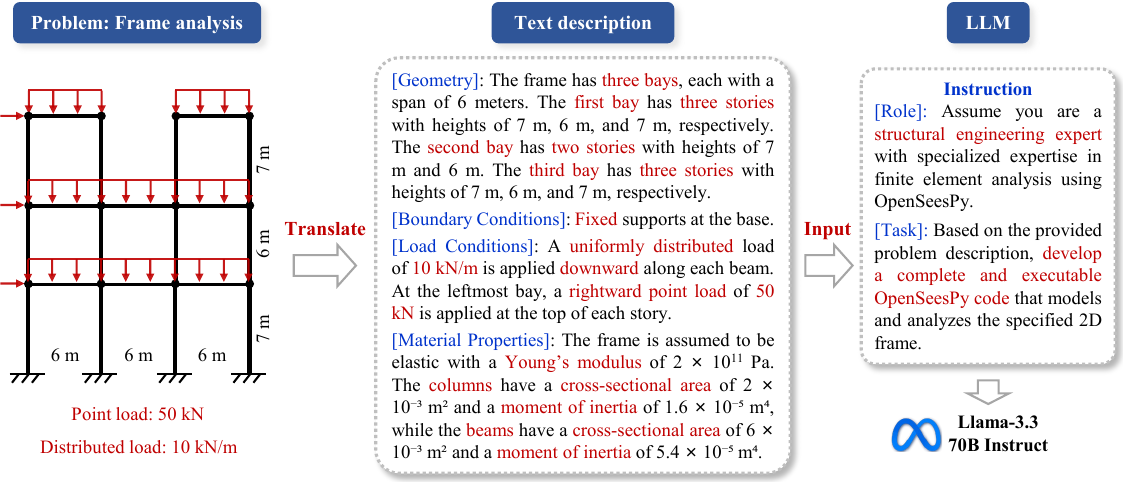}
\caption{Workflow to assess the performance of Llama model in 2D frame structural analysis.}
\label{Figure2}
\end{figure*}

The workflow for evaluating the performance of Llama-3.3 70B Instruct model is illustrated in \cref{Figure2}, where the LLM is prompted to generate OpenSeesPy code for 2D frame structural analysis. The evaluation process begins by constructing a structured text description for each frame problem. This step follows the approach outlined in \cite{wan2025som}, which specifies frame geometry, boundary conditions, load conditions, and material properties. The description is then supplemented with two types of instructions to guide the model’s generative behavior: (a) a role instruction that assigns the LLM the role of a structural engineering expert with specialized expertise in FEM analysis using OpenSeesPy, and (b) a task instruction that directs the LLM to produce complete and executable code for modeling and analyzing the specified frame. The problem description and instructions are provided as input to the Llama-3.3 70B Instruct model. Given the stochasticity of the LLM outputs, each problem is evaluated across ten independent trials. The performance is quantified by accuracy, defined as the proportion of correctly generated codes relative to the total number of runs.

\begin{figure*}[htbp]
\centering
\includegraphics[width=0.9\textwidth]{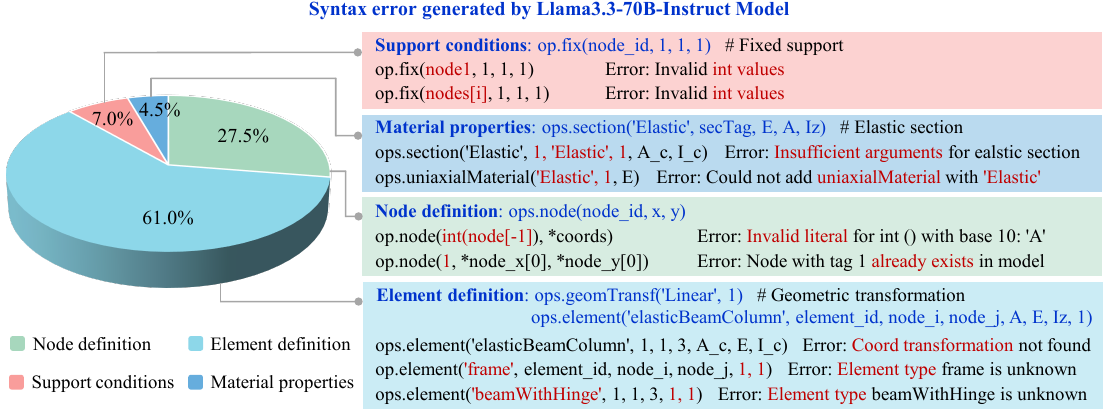}
\caption{Syntax errors produced by the Llama model in frame structural analysis.}
\label{Figure3}
\end{figure*}

The evaluation reveals a significant deficiency in the Llama-3.3 70B Instruct model to perform frame structural analysis. Across all 20 problems, the LLM consistently fails to generate executable code: each of the ten runs terminate with syntax errors. \cref{Figure3} summarizes the distribution of error sources: element definition (61.0\%), node definition (27.5\%), support conditions (7.0\%), and material properties (4.5\%). Specifically, element definition errors are the most prevalent, arising primarily from two issues: (a) omission of coordinate definitions, which are prerequisites for creating beam-column elements, and (b) use of non-existent element types. The former reflects weak adherence to procedural dependencies, whereas the latter suggests conflation among command syntaxes across FEM packages. Node definition exhibits two dominant failure modes: (a) assignment of invalid data types, where non-integer values such as character literals are used as node IDs, and (b) creation of duplicate node IDs within the model. These failures indicate weak enforcement of data types and inadequate consistency checks during code generation. Errors in support conditions largely mirror those in node definitions, resulting from incorrect data types for node IDs. Material property errors involve missing or conflicting arguments in elastic section definitions, reflecting the model’s inability to comply with strict syntactic requirements and to maintain logical consistency. Overall, these results demonstrate the LLM is prone to hallucination in code generation for frame structural analysis, lacking the precision, consistency, and reliability required for practical engineering applications.

\begin{figure*}[htbp]
\centering
\includegraphics[width=0.9\textwidth]{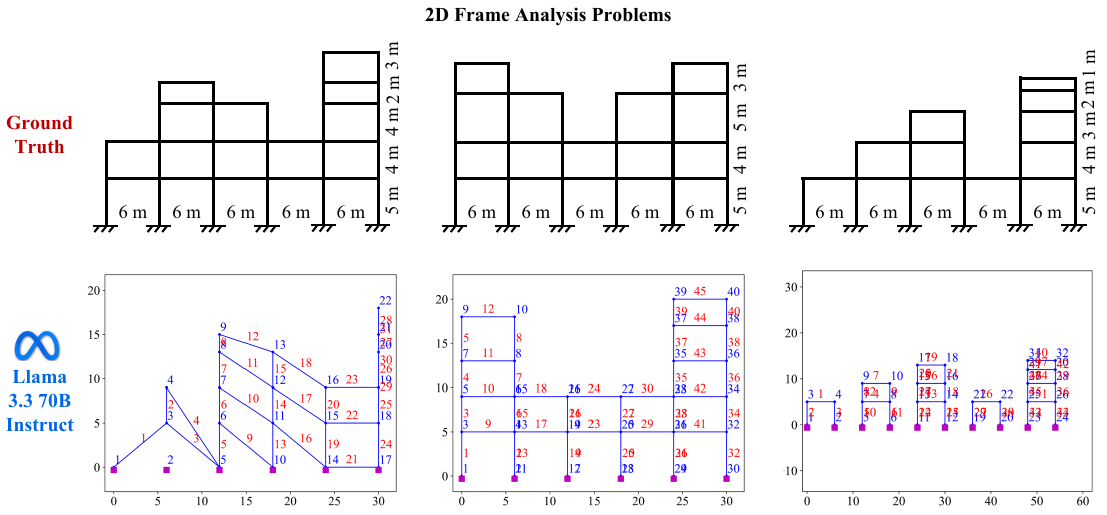}
\caption{Evaluation of Llama model’s spatial reasoning capability through frame geometry generation.}
\label{Figure4}
\end{figure*}

In addition to syntax error analysis, the spatial reasoning capability of the Llama-3.3 70B Instruct model is evaluated through its generation of frame geometries. To enable this evaluation, syntax errors in node and element definitions are manually corrected, allowing visualization of the geometric outputs. The results, illustrated by three representative cases in \cref{Figure4}, reveal profound deficiencies in the LLM’s ability to construct coherent 2D frame geometries. In case 1, the LLM demonstrates a fundamental misunderstanding of structural topology: instead of producing an orthogonal grid of beams and columns, it creates a truss-like configuration with diagonal members. It also omits nodes in the fifth bay where story heights change, reflecting an inability to preserve structural integrity under topological variations. Cases 2 and 3 highlight failures in structural connectivity. In case 2, while the LLM produces a frame-like geometry, it redundantly defines duplicate nodes and elements at the intersections of adjacent bays. This issue is even more significant in case 3, where the LLM treats each bay as an isolated entity, modeling them separately with no shared nodes or elements. These behaviors indicate that the LLM fails to capture the fundamental engineering principle of employing shared nodes and elements to ensure continuity across structural components. Collectively, these findings underscore spatial reasoning as a critical challenge for applying LLMs to frame structural analysis.

\section{A large language model-based multi-agent system for frame structural analysis}
\label{sec:others}

This section introduces the multi-agent system for FEM of 2D frames. In subsection \ref{sec:overall_system}, the overall system design is discussed. Specifically, the system includes problem analysis agent, geometry agent, code translation agent, model validation agent, and load agent. Among the five agents, the gemometry agent and the model validation agent have a more complicated architecture. Therefore, detailed discussions are presented in subsections \ref{sec:geometry_agent} and \ref{sec:validation_agent}.

\subsection{Overall system design}
\label{sec:overall_system}

\begin{figure*}[htbp]
\centering
\includegraphics[width=0.9\textwidth]{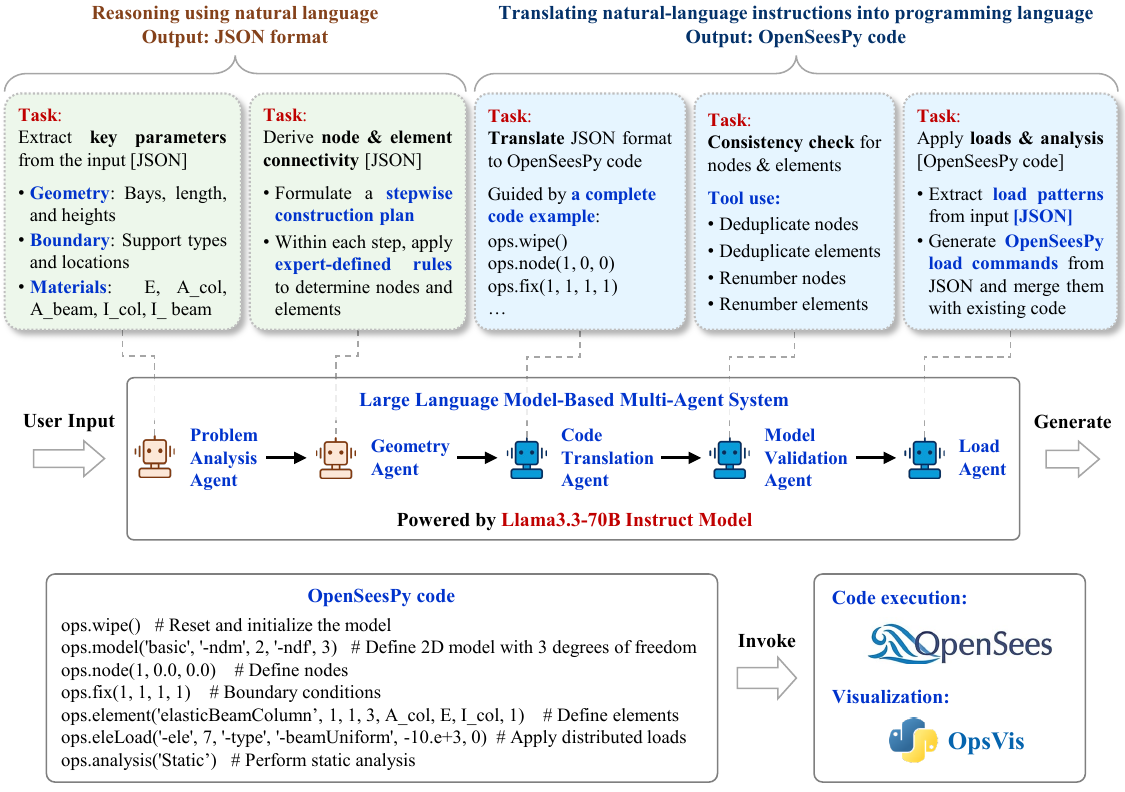}
\caption{System design of the LLM-based multi-agent framework for automated frame structural analysis.}
\label{Figure5}
\end{figure*}

Given the inherent complexity of frame structural analysis, the proposed system adopts a multi-agent architecture that decomposes the overarching task into a sequence of manageable subtasks, as illustrated in \cref{Figure5}. Each subtask is managed by a specialized agent powered by the Llama-3 70B Instruct model, enabling modular task execution within an integrated framework. Specifically, the workflow begins with a natural language description of the frame problem provided by the user. This input is sequentially processed by a pipeline of five agents: problem analysis agent, geometry agent, code translation agent, model validation agent, and load agent. The pipeline generates OpenSeesPy code for modeling and analyzing the specified 2D frame structure. Then, the system invokes the OpenSeesPy library to execute the code and employ OpsVis to visualize the structural model and analysis results. This design provides an end-to-end automated workflow that transforms natural language problem descriptions into executable and visualized structural analysis, thereby minimizing manual intervention and enhancing FEM efficiency. 

The role of each agent within the proposed system is defined as follows. The problem analysis agent acts as the initial interpreter, parsing the user’s input to extract geometry, boundary, and material parameters. Specifically, geometry parameters include the number of bays, bay spans, number of stories, and story heights. Boundary conditions specify support types and locations, while material parameters comprise Young’s modulus, cross-sectional areas, and moments of inertia of beams and columns. These parameters are organized into a structured JSON object for downstream processing, as illustrated in the left column of Table~\ref{tab:json}. The geometry agent then formulates a stepwise construction plan by assembling the structure bay by bay and story by story. At each step, expert-defined rules are applied to derive node coordinates and element connectivity, producing a JSON representation of the structural topology, as shown in the right column of Table~\ref{tab:json}. The detailed architecture of the geometry agent will be presented in Subsection 3.2. 

\begin{table}[htbp]
\centering
\captionsetup{skip=5pt}
\caption{Representative JSON outputs from the problem analysis and geometry agents.}
\label{tab:json}
\begin{tabular}{p{3.8cm} p{8.8cm}}
\toprule
Problem analysis agent &
Geometry agent 
\\
\midrule
\ttfamily\scriptsize
\begin{minipage}[c]{\linewidth}
\begin{verbatim}
{
  "Total_bays": 5,
  "Geometry": {
    "Bay": 1,
    "Span": 6,
    "Total_stories": 3,
    "Heights": [5, 4, 5]
  },
  // ... Omitted
  "Supports": "Fixed",
  "Material": {
    "E": 2e11,
    "A_col": 2e-3,
    "A_gir": 6e-3,
    "I_col": 1.6e-5,
    "I_gir": 5.4e-5
  }
}
\end{verbatim}
\end{minipage}
&
\ttfamily\scriptsize
\begin{minipage}[c]{\linewidth}
\begin{verbatim}
{
  "Construction_steps": [
    {
      "Step": 1,
      "Bay": 1,
      "Story": 1,
      "Rule": "Rule_1",
      "Nodes": [
        {"ID": 1, "x": 0, "y": 0, "Desc": "Bottom left"},
        {"ID": 2, "x": 6, "y": 0, "Desc": "Bottom right"},
        {"ID": 3, "x": 0, "y": 5, "Desc": "Top left"},
        {"ID": 4, "x": 6, "y": 5, "Desc": "Top right"}
      ],
      "Elements": [
        {"ID": 1, "Coord": [(0,0), (0,5)], "Desc": "Left column"},
        {"ID": 2, "Coord": [(6,0), (6,5)], "Desc": "Right column"},
        {"ID": 3, "Coord": [(0,5), (6,5)], "Desc": "Top girder"}
      ],
      "Boundary_conditions": [
        {"Node_ID": 1, "Constraints": "Fixed"},
        {"Node_ID": 2, "Constraints": "Fixed"}
      ]
    }
    // ... Omitted
  ]
}
\end{verbatim}
\end{minipage}
\\
\bottomrule
\end{tabular}
\end{table}

The code translation agent converts the structured JSON specification of the frame geometry into executable OpenSeesPy code. This is achieved by providing a complete code example and instructing the LLM to follow the prescribed procedure and syntax. A representative example of the code produced by the translation agent is provided in \hyperref[appendix:translation and validation]{Appendix A}. The model validation agent then refines the translated code by invoking a suite of predefined tools to perform consistency checks, including the removal of duplicate nodes or elements and the renumbering of IDs. The detailed architecture of the model validation agent will be presented in Subsection 3.3, and a revised code example is also included in \hyperref[appendix:translation and validation]{Appendix A}. Finally, the load agent extracts load patterns from the user’s input and generates the corresponding OpenSeesPy commands, which are integrated with the validated code to produce the complete structural analysis program. An illustrative example of the final output of the multi-agent system is provided in \hyperref[appendix:complete]{Appendix B}.

A central design principle of the multi-agent system is the decoupling of geometric reasoning from code generation, implemented through a two-stage pipeline, as shown in \cref{Figure5}. In the first stage, the problem analysis and geometry agents operate in natural language to interpret the user’s input and infer the structural topology. Instead of directly generating codes, this stage outputs a structured JSON representation that explicitly encodes the construction logic of the geometric model. The second stage translates these structured specifications into executable OpenSeesPy code, achieved through the collaborative efforts of the code translation, model validation, and load agents. This two-stage design addresses two inherent challenges of LLMs separately: (a) limited spatial reasoning and (b) instability in code generation. For the former, the system leverages the LLM’s strength in linguistic reasoning, enabling it to resolve topological relationships using natural language before transitioning to code. For the latter, the LLM assumes a constrained translation role, following the predefined geometry in the JSON specification to generate code. By assigning each agent a specialized and well-defined scope, the system reduces ambiguity, improves robustness, and minimizes the risk of hallucinations.

\subsection{Architecture of the geometry agent}
\label{sec:geometry_agent}

The geometry agent is responsible for translating high-level, abstract descriptions of the frame into a precise geometric model defined by node coordinates and element connectivity. To achieve this, the agent is designed to emulate the systematic and incremental workflow of a human engineer when assembling structural models. This design reframes the complex spatial reasoning task into a structured, step-by-step process, enabling the agent to focus on resolving local geometric relationships incrementally. By reducing the agent’s cognitive burden, this approach is expected to enhance the consistency and accuracy of the constructed topology. Specifically, the workflow comprises two stages: first, the assembly sequence is planned at a conceptual level, and second, the nodes and elements are instantiated within each construction step under the guidance of explicit and expert-defined rules.

\begin{figure*}[htbp]
\centering
\includegraphics[width=0.9\textwidth]{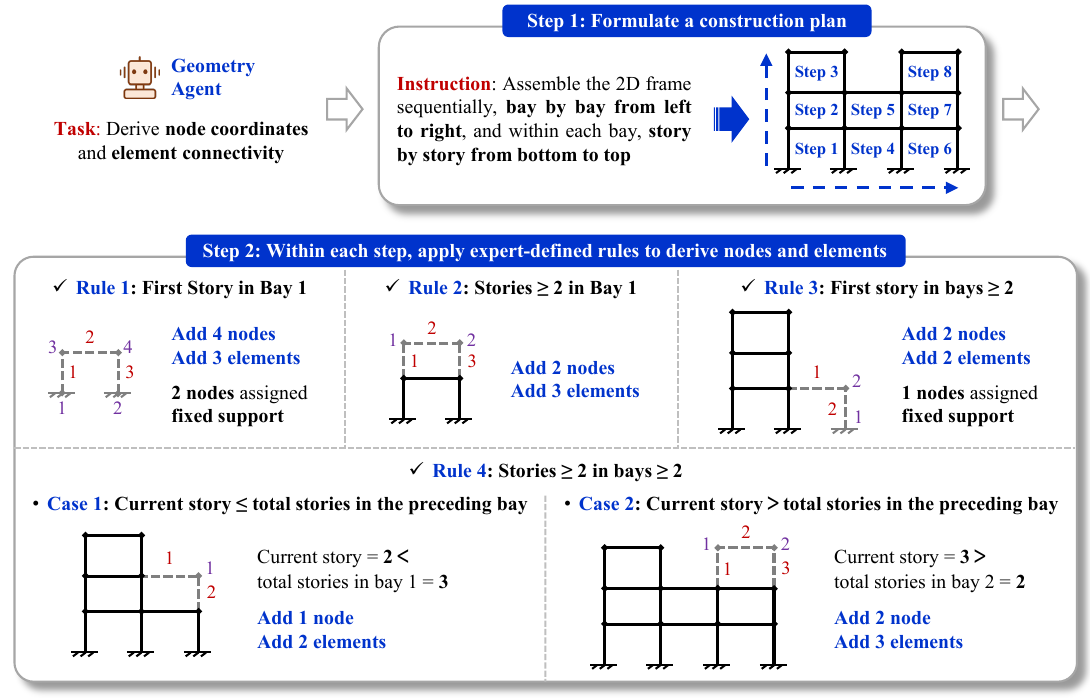}
\caption{Workflow of the geometry agent for deriving node coordinates and element connectivity.}
\label{Figure6}
\end{figure*}

In the first stage, the geometry agent mirrors human modeling practices to formulate a stepwise construction plan for the 2D frame. To reduce the risk of hallucinations in LLMs, a deterministic modeling order is explicitly specified: the frame is assembled sequentially, bay by bay from left to right, and within each bay, story by story from bottom to top. This order constrains the search space and ensures that the topology is generated in a controlled and consistent manner. As illustrated in \cref{Figure6} using a 3‑2‑3 frame example, this planning approach effectively transforms the complex 2D spatial reasoning problem into a sequence of discrete construction steps, laying a robust foundation for deriving nodes and elements in the subsequent stage.

Building on the construction plan, the second stage applies a set of expert-defined rules to derive node coordinates and element connectivity within each step. These rules formalize the required spatial logic for frame modeling, specifying how nodes and elements are added under different topological scenarios, as illustrated in \cref{Figure6}:
\begin{enumerate}[label=\textbullet~Rule \arabic*:, leftmargin=*, itemsep=4pt]
  \item For the first story in the first bay, define four nodes and three elements,
        and assign fixed supports to the two base nodes.
  \item For each additional story ($\geq 2$) in the first bay, add two new nodes
        and three new elements.
  \item For the first story of any subsequent bay ($\geq 2$), add two new nodes and
        two new elements, and assign a fixed support to the new base node.
  \item For each additional story ($\geq 2$) in a subsequent bay ($\geq 2$), perform
        a conditional judgment: if the current story is less than or equal to the
        total number of stories in the preceding bay, add one new node and two new
        elements; otherwise, add two new nodes and three new elements.
\end{enumerate}

These rules are embedded into the system prompt of the LLM, encoding the the implicit logic of human engineers into the agent’s workflow. By adopting these rules, the stochastic generative behavior of the LLM is transformed into a transparent and interpretable reasoning process, where each decision can be traced to a clearly defined rule and construction step. This design enhances traceability, reduces ambiguity, and improves consistency in the generation of structural topologies.

\subsection{Architecture of the model validation agent}
\label{sec:validation_agent}

\begin{figure*}[htbp]
\centering
\includegraphics[width=0.9\textwidth]{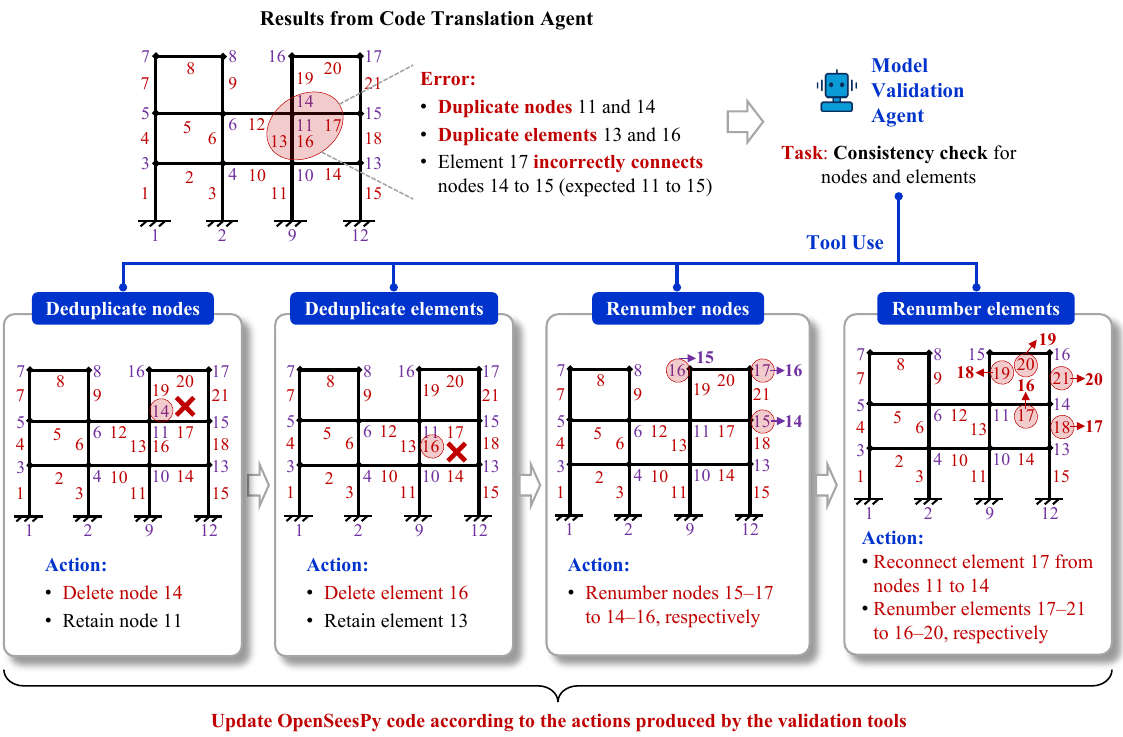}
\caption{Workflow of the model validation agent for code consistency checks.}
\label{Figure7}
\end{figure*}

Although the expert-defined rules embedded within the geometry agent improve the accuracy of node and element definitions, LLMs remain susceptible to hallucinations, often generating redundant or inconsistent outputs such as duplicate nodes, duplicate elements, or incorrect connectivity. To mitigate these issues, a model validation agent is introduced to perform consistency checks of the codes. This agent adopts a deterministic, tool-based validation framework, in which Python functions are invoked by the LLM to ensure rigorous and reliable verification. The workflow of the model validation agent is illustrated in \cref{Figure7}. It receives the OpenSeesPy code produced by the code translation agent and sequentially executes four specialized tools to identify and correct: (a) duplicate nodes, (b) duplicate elements, (c) irregular node numbering, and (d) inconsistent element numbering. Each tool outputs a set of corrective actions expressed in natural language, which are then interpreted by the agent to refine the code accordingly.

\cref{Figure7} illustrates the functionality of each tool within the model validation agent using a representative example. The initial OpenSeesPy code contains several inconsistencies, including duplicate nodes (nodes 11 and 14), duplicate elements (elements 13 and 16), and an incorrectly connected element (element 17). Particularly, element 17 is erroneously connected between nodes 14 and 15, rather than the intended nodes 11 and 15. These errors are corrected through a sequence of validation steps. First, the node deduplication tool identifies and removes redundant nodes, retaining the one with the smaller ID (node 11) and deleting its duplicate (node 14). Second, the element deduplication tool applies the same principle, preserving the lower-indexed element (element 13) and removing its duplicate (element 16). Third, the node renumbering tool reassigns node identifiers to maintain sequential order, updating nodes 15–17 to 14–16. Finally, the element renumbering tool performs two operations: it first corrects the connectivity of elements affected by node deletion. Herein, since element 17 is originally connected to the deleted node 14, the tool outputs an action to reconnect it to the retained node 11. It then reindexes elements 17–21 to 16–20 to ensure contiguous numbering. Through this multi-step verification workflow, the model validation agent ensures that the revised OpenSeesPy code is logically consistent and structurally valid.

\section{Results and discussion}

\subsection{Performance of the LLM-based multi-agent system}

\begin{figure*}[htbp]
\centering
\includegraphics[width=0.85\textwidth]{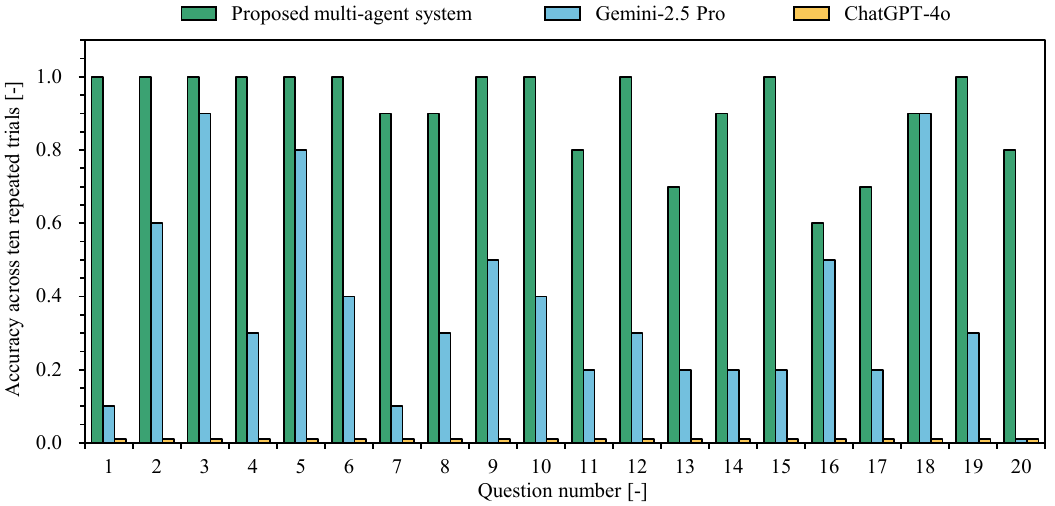}
\caption{Performance of the LLM-based multi-agent system on the benchmark dataset.}
\label{Figure8}
\end{figure*}

The performance of the proposed LLM-based multi-agent system is evaluated using the benchmark dataset comprising 20 frame analysis problems. To assess the consistency and robustness of the generated codes, each problem is tested across ten independent runs. As shown in \cref{Figure8}, the proposed system exhibits reliable and stable performance, achieving over 80\% accuracy in most cases. This highlights the system’s capability to address spatial reasoning and syntactic consistency challenges that caused failures in the vanilla Llama model. Specifically, in simple scenarios, such as three-bay frame structures, the system achieves perfect performance across all trials, consistently generating error-free OpenSeesPy codes. For complex topologies, such as five-bay frames, the system maintains strong performance, attaining an average accuracy of 88\% across 15 problems. These results confirm the system’s ability to follow expert-defined rules even under irregular geometric configurations, emphasizing its versatility and generalizability. However, a modest performance decline is observed in specific cases. The accuracy drops to 70\% for the 2-4-3-2-5 and 2-3-3-2-5 frames, and to 60\% for the 2-3-1-4-5 frame. These failures suggest that the system remains susceptible to hallucinations in scenarios involving lengthy sequential steps and frequent conditional rule applications.

\begin{figure*}[htbp]
\centering
\includegraphics[width=0.9\textwidth]{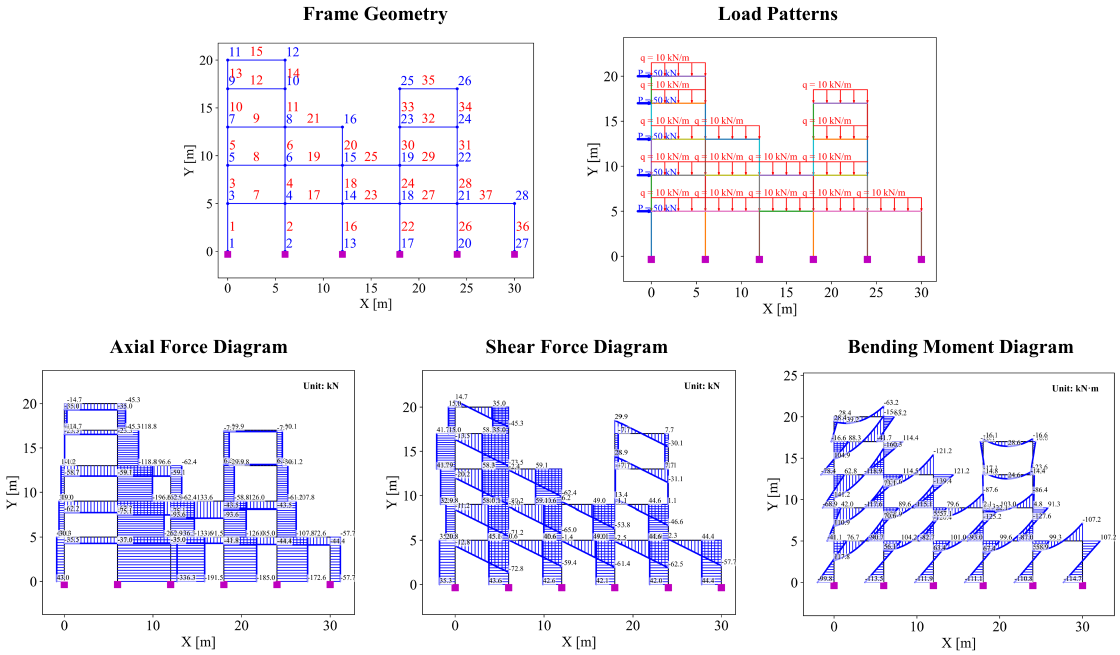}
\caption{Visualization of the proposed multi-agent system for 2D frame structural analysis.}
\label{Figure9}
\end{figure*}

In addition to automated code generation, the proposed multi-agent system integrates the OpsVis package \citep{kokot_opsvis_2024} to visualize the structural model and analysis results, as depicted in \cref{Figure9}. Using a representative 5-3-2-4-1 frame structure, the system produces five graphical outputs: frame geometry, load patterns, and the distributions of axial force, shear force, and bending moment. Specifically, the geometry diagram provides a clear depiction of nodes, elements, and boundary conditions, while the load diagram illustrates the magnitude and direction of applied point and distributed loads. The internal force diagrams offer detailed insights into the axial, shear, and bending responses of individual members, capturing the mechanical behavior of the frame under loading. Collectively, these visualizations provide human engineers with an intuitive interface to verify the correctness of the generated codes. Moreover, they serve as diagnostic tools by exposing inconsistencies in the structural response, facilitating the rapid detection and correction of modeling errors. Overall, this visualization enables an effective human-in-the-loop verification process, improving the interpretability, reliability, and practical utility of the proposed multi-agent system.

\subsection{Comparison with state-of-the-art LLMs}

To further validate the effectiveness of the proposed multi-agent system, a comparative analysis is conducted against two state-of-the-art (SOTA) LLMs: Gemini 2.5 Pro \citep{deepmind2025gemini} and ChatGPT 4o \citep{openai2024gpt4o}. Both models are evaluated using the benchmark dataset described in Section 2.1. They receive the same textual prompts as the Llama model to generate OpenSeesPy code for analyzing the mechanical behavior of frame structures. The results, presented in \cref{Figure8}, reveal that proposed system consistently outperforms both SOTA models across all 20 benchmark problems. Specifically, the Gemini model achieves an average accuracy of 37\%. Notably, its performance demonstrates significant variability, with accuracy ranging from as high as 90\% to as low as 0\%. The ChatGPT model, despite its reputation as a leading general-purpose LLM, fails to generate any valid OpenSeesPy code across all ten trials for each of the 20 problems, resulting in a consistent accuracy of 0\%. These results underscore the limitations of general-purpose LLMs in specialized engineering tasks and demonstrate the need of domain-specific LLMs.

\begin{figure*}[htbp]
\centering
\includegraphics[width=0.9\textwidth]{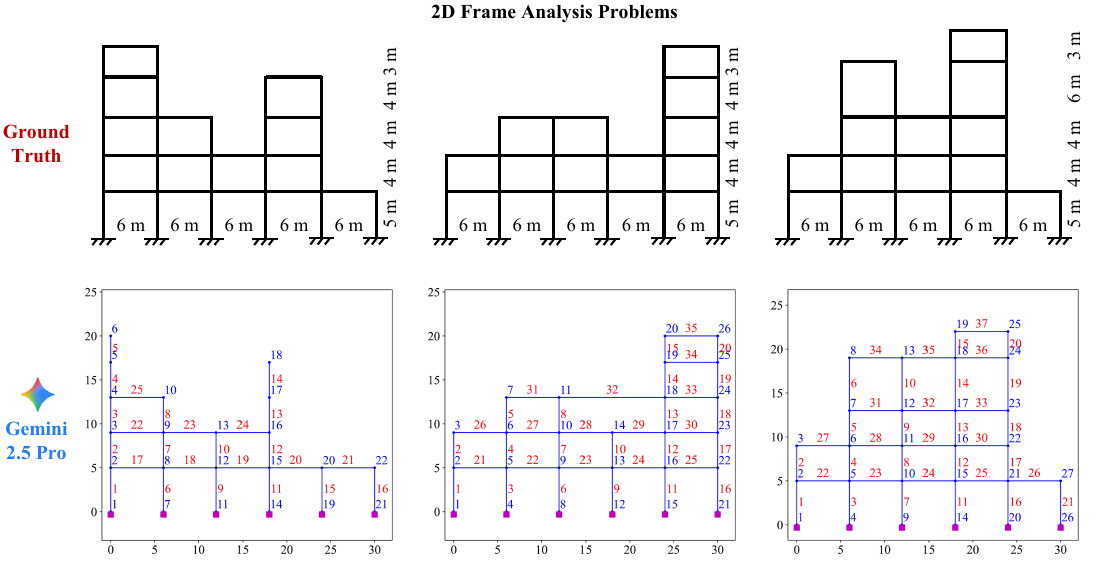}
\caption{Representative errors produced by Gemini model in generating 2D frame geometry.}
\label{Figure10}
\end{figure*}

A detailed examination is conducted to elucidate the sources of failure in the Gemini model. Unlike the Llama model, Gemini demonstrates a solid command of OpenSeesPy syntax: among 200 runs (20 problems with 10 trials each), 88.5\% of the generated code executes without runtime errors. Despite this high execution rate, most errors stem from incorrect geometric construction of frame structures. \cref{Figure10} shows several representative failure scenarios. The Gemini model generally maintains orthogonality, producing square or rectangular grids and avoiding diagonal element connections. This indicates that the Gemini model demonstrates a basic understanding of frame structure topologies, outperforming the Llama model in capturing fundamental geometric principles. However, critical topological flaws persist. For the 5-3-2-4-1 frame, the model omits necessary nodes to complete a story within one bay, resulting in an incomplete structure. For the 2-3-3-2-5 frame, the model cross-connects nodes between nonadjacent bays (from the third to the fifth bay), producing an invalid story configuration. For the 2-4-3-5-1 frame, an incorrect beam placement in the third bay introduces a redundant story. These errors reveal that while the Gemini model exhibits syntactic proficiency, it lacks robust spatial reasoning capabilities required for generating topologically valid and mechanically sound structural models.

\begin{figure*}[htbp]
\centering
\includegraphics[width=0.9\textwidth]{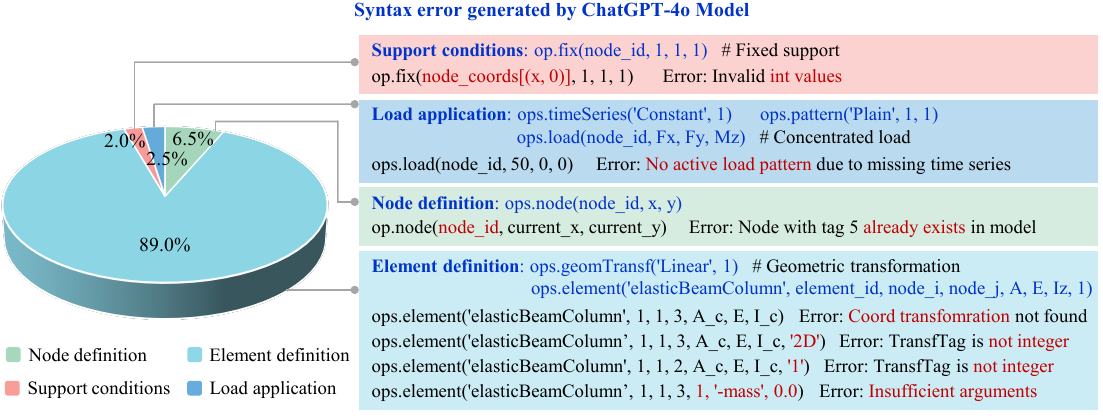}
\caption{Syntax errors produced by the ChatGPT model in frame structural analysis.}
\label{Figure11}
\end{figure*}

ChatGPT is worse than Gemini because it has runtime errors (i.e., the generated codes cannot run). \cref{Figure11} show the distribution of these runtime errors. The vast majority of errors (89.0\%) occur during element definition. These syntactical violations include: (a) omission of coordinate definitions, which is also a major source of failure in the Llama model; (b) incorrect data types assigned to coordinate transformation tags, reflecting a failure to distinguish between integer identifiers and string labels; and (c) incomplete specification of required arguments, leading to invalid element declarations. Errors related to node definitions account for 6.5\% of failures, primarily due to duplicate node IDs. Load application errors contribute 2.5\%, largely resulting from the omission of time series and load pattern definitions, both of which are mandatory to activate load commands. The remaining 2.0\% of errors are associated with support condition definitions, where inappropriate data types are assigned to node identifiers. Collectively, these errors reveal that the ChatGPT-4o model lacks the syntactic knowledge of domain-specific programming frameworks such as OpenSeesPy, thereby limiting its applicability to structural engineering tasks.

\begin{figure*}[htbp]
\centering
\includegraphics[width=0.8\textwidth]{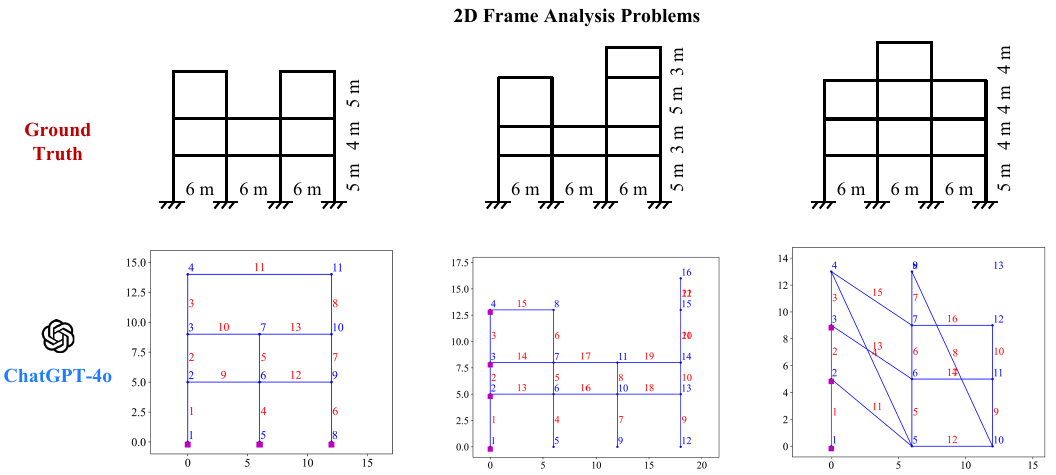}
\caption{Representative errors produced by ChatGPT model in generating 2D frame geometry.}
\label{Figure12}
\end{figure*}

Additionally, the spatial reasoning capabilities of ChatGPT model is examined. The syntactic errors are manually fixed and then the geometric configurations produced by ChatGPT are examined, which are shown in \cref{Figure12}. These cases involve relatively simple three-bay frames with topologies of 3-2-3, 3-2-4, and 3-4-3. Despite their simplicity, the ChatGPT model consistently fails to generate correct structural geometries. Specifically, for the 3-2-3 frame, ChatGPT fails to understand the overall bay number and produces a structure with only two bays. For the 3-2-4 frame, ChatGPT neglects certain nodes required for constructing a complete story, leading to an incomplete structure layout. The most severe errors are observed in the 3-4-3 frame, where ChatGPT introduces diagonal elements between non-adjacent nodes, demonstrating a fundamental misunderstanding of basic frame topology. In summary, the failures of SOTA models highlight the inherent limitations of general-purpose LLMs in addressing domain-specific challenges in structural engineering, particularly in spatial reasoning and code consistency. These findings validate the core design philosophy of the proposed multi-agent system, which decomposes these tasks and delegates them to specialized agents.

\subsection{Ablation experiment: Validity of task decomposition}

\begin{figure*}[htbp]
\centering
\includegraphics[width=0.9\textwidth]{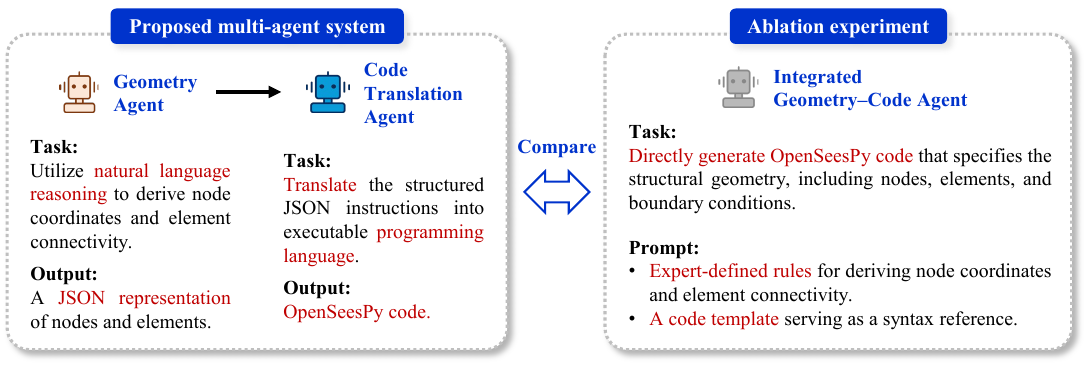}
\caption{Ablation experiment illustrating the role of decomposing geometric reasoning and code translation.}
\label{Figure13}
\end{figure*}

One core design principle of the proposed multi-agent system is the decomposition of geometric reasoning and code generation into two distinct tasks, managed by the geometry agent and the code translation agent, respectively. To validate the effectiveness of this modular architecture, an ablation experiment is conducted in which these two tasks are integrated into a single geometry–code agent, as illustrated in \cref{Figure13}. This integrated agent is tasked with directly generating OpenSeesPy code that defines the structural geometry, including node coordinates, element connectivity, and boundary conditions. For a fair comparison, the integrated agent is provided with the same prompts as the original system: a set of expert-defined rules for determining geometric configurations and a complete code template to serve as a syntax reference.

The ablation experiment is conducted on three representative frame structures, with each case tested over ten independent runs. The results, summarized in Table~\ref{tab:ablation_accuracy}, reveal a stark contrast in performance. The proposed multi-agent system consistently achieves 100\% accuracy across all cases, while the integrated geometry–code agent fails completely and has an accuracy of 0\%. Notably, the code generated by the integrated agent is syntactically correct and executable, suggesting that it references the code template to adhere to syntax rules. However, the failures arise from the agent’s inability to correctly apply the expert-defined rules governing geometric construction. A typical error occurs when the current story exceeds the total height of the adjacent bay. In such cases, the integrated agent omits the top-left node required to complete the story, leading to incomplete or topologically inconsistent frame structures.

\begin{table}[htbp]
  \centering
  \captionsetup{skip=5pt}
  \caption{Performance comparison between the multi-agent system and integrated geometry-code agent.}
  \renewcommand{\arraystretch}{1.2}
  \begin{tabular}{l S[table-format=1.2] S[table-format=1.2] S[table-format=1.2]}
    \toprule
    \diagbox[width=13.7em, height=3.5ex, trim=l]{Model}{Accuracy} & {Frame: 3-2-3} & {Frame: 3-2-4} & {Frame: 3-4-3} \\
    \midrule
    Proposed Multi-agent System      & 1.00 & 1.00 & 1.00 \\
    Integrated Geometry--Code Agent  & 0.00 & 0.00 & 0.00 \\
    \bottomrule
  \end{tabular}
  \label{tab:ablation_accuracy}
\end{table}

These findings yield two critical insights. First, they demonstrate that even when provided with expert-defined rules and a syntactic template, LLMs struggle to manage spatial reasoning and code generation tasks concurrently. The coupling of these tasks appears to exceed the reasoning capacity of current LLMs, resulting in structurally invalid outputs despite syntactic correctness. Second, the failure of the integrated agent highlights the validity of the proposed multi-agent system. That is by decoupling geometric reasoning from code translation, the multi-agent system assigns each task to a specialized agent with a clearly defined scope. This modular system design reduces ambiguity, enhances reliability, and promotes logical consistency in the generated structural models.

\subsection{Runtime and costs}

\begin{figure*}[htbp]
\centering
\includegraphics[width=0.45\textwidth]{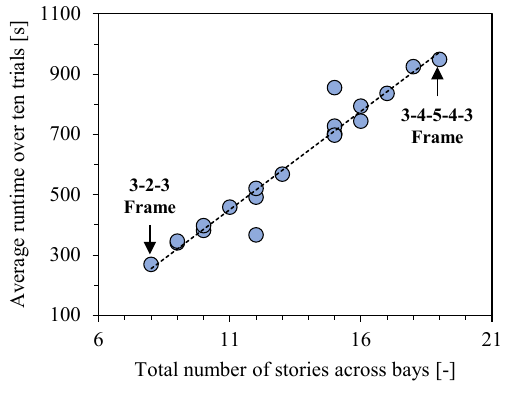}
\caption{Average runtime across ten trails for each frame structure in the benchmark dataset.}
\label{Figure14}
\end{figure*}

Beyond accuracy, the practicality of the proposed multi-agent system is assessed in terms of computational efficiency. For each of the 20 frame structures in the benchmark dataset, the average runtime over ten trials is calculated, as illustrated in \cref{Figure14}. The results reveal that the runtime increases as the structural complexity increases. The 3-2-3 frame requires the least time of 269.2 seconds, whereas the 3-4-5-4-3 frame demands the most time of 949.0 seconds to generate the OpenSeesPy code. This trend reflects that the increased topological complexity involves a greater number of construction steps, thereby increasing the computational load. To demonstrate the efficiency gains, the system’s performance is compared with manual implementation. One of the authors manually codes the 3-2-3 frame using OpenSeesPy, which takes 17 minutes and 24 seconds. This comparison indicates that the proposed multi-agent system can significantly improve the workflow efficiency in frame structural modeling.

An additional observation is that the runtime of the proposed multi-agent system exhibits an approximately linear relationship with the total number of stories across bays, as illustrated in \cref{Figure14}. This behavior is a direct consequence of the incremental construction strategy embedded in the geometry agent. At each step, the system identifies the current construction location and applies the corresponding expert-defined rule, ensuring that the computational effort per step remains relatively constant. In contrast, for human engineers, the difficulty of manual modeling tends to increase nonlinearly with structural complexity. This is because managing node identifiers and element connectivity becomes increasingly challenging with additional stories. The system’s ability to maintain linear runtime growth, regardless of geometric irregularity, demonstrates its scalability and highlights its potential for application to large-scale and complex structural systems.

Another key aspect of practical deployment of the proposed multi-agent system is its economic cost. To this end, a comparative analysis is conducted between the costs of the proposed multi-agent system and SOTA LLMs. Table~\ref{tab:running_costs} summarizes the runtime across the benchmark datasets. The comparison encompasses input and output token consumption, API pricing, and the resulting total cost. Specifically, SOTA models consume relatively few input tokens as they directly take the user’s problem description as input. In contrast, the proposed system exhibits higher input token usage due to its sequential multi-agent architecture, wherein outputs from intermediate agents are recursively passed as inputs to downstream agents. Regarding output token usage, the Gemini model generates more tokens than ChatGPT, indicating that it performs more elaborate reasoning steps that may contribute to its relatively higher accuracy. In contrast, the proposed multi-agent system demonstrates higher output token consumption, attributable to its multi-stage reasoning process.

\begin{table}[htbp]
  \centering
  \captionsetup{skip=5pt}
  \caption{Comparison of running costs between the proposed multi-agent system and state-of-the-art LLMs.}
  \begin{threeparttable}
  \renewcommand{\arraystretch}{1.2}
  \begin{tabular}{
    p{2.7cm}  
    p{2.3cm}  
    p{2cm}  
    p{2.3cm}  
    p{2.1cm}  
    p{2.8cm}  
  }
    \toprule
    Model & Input Tokens & Input Pricing & Output Tokens & Output Pricing & Total Running Cost \\
    \midrule
    Proposed system
        & 23,110 -- 56,951 & \$0.12 / 1M\tnote{*}  
        & 15,540 -- 53,141 & \$0.30 / 1M\tnote{*}  
        & \$0.0074 -- \$0.0228 \\

    ChatGPT-4o 
        & 345 -- 467 & \$2.50 / 1M 
        & 934 -- 1,290 & \$10.00 / 1M 
        & \$0.0104 -- \$0.0138 \\

    Gemini 2.5 Pro 
        & 345 -- 467 & \$1.25 / 1M 
        & 2,305 -- 4,402 & \$10.00 / 1M 
        & \$0.0236 -- \$0.0447 \\
    \bottomrule
  \end{tabular}
  \begin{tablenotes}
    \item[*] Token pricing for the proposed system is based on the Llama 3.3 70B Instruct model via the Lambda API.
  \end{tablenotes}
  \end{threeparttable}
  \label{tab:running_costs}
\end{table}

Despite the increased token consumption, the overall running cost of the proposed system remains highly competitive, primarily due to the low pricing of the Llama 3.3 70B Instruct model that underpins its architecture. Specifically, the token cost of \$0.12 per million tokens for inputs and \$0.30 per million tokens for outputs. These costs are one to two orders of magnitude lower than that of ChatGPT-4o (\$2.50/\$10.00) and Gemini 2.5 Pro (\$1.25/\$10.00). This results in the lower total running cost of the proposed system  than that of Gemini and only slightly higher than that of ChatGPT. Notably, this modest cost increase relative to ChatGPT yields a substantial performance improvement, while Gemini’s marginal accuracy gains are accompanied by a higher financial burden. These results highlight that the proposed multi-agent system offers a cost-effective yet high-performing solution to automate frame structural analysis. By integrating a lightweight and economically efficient LLM within a rule-guided multi-agent framework, the system delivers practical scalability without incurring the prohibitive costs associated with SOTA LLMs. This balance between cost and performance highlights the proposed system’s potential for real-world engineering applications.

\section{Summary and conclusions}

This paper develops a large language model (LLM)-based multi-agent system for automated structural analysis of 2D frames, aiming to address the limitations of spatial reasoning and syntax consistencies in state-of-the-art LLMs. The proposed system adopts a task decomposition strategy, partitioning the structural analysis task into a sequence of specialized subtasks. These subtasks are assigned to five dedicated agents powered by the Llama-3 70B Instruct model: a problem analysis agent, a geometry agent, a code translation agent, a model validation agent, and a load agent. A central design philosophy of the system lies in the decoupling of geometric reasoning from code generation. Specifically, the problem analysis and geometry agents operate in natural language to interpret user input and derive the structural topology, while the subsequent agents translate these structured representations into executable OpenSeesPy code. Within this framework, the geometry agent plays a pivotal role, which emulates the incremental construction workflow of human engineers. Expert-defined rules are embedded within the agent to guide the step-by-step construction of nodes and elements. This reduces the ambiguity and enhances consistency in geometry generation. Experimental evaluations on a benchmark dataset of 20 frame structures are performed and the following conclusions are drawn.

\begin{itemize}
    \item The proposed multi-agent system achieves accuracy exceeding 80\% in most cases across ten repeated trials, consistently generating valid structural analysis code even for irregular geometries. It also significantly outperforms state-of-the-art LLMs (i.e., Gemini 2.5 Pro and ChatGPT 4o). These results demonstrate the reliability and robustness of the proposed system, representing a step forward in developing automated tools for structural engineering practice.

    \item The ablation experiments demonstrate the critical role of decomposing geometric reasoning and code generation in improving system performance. When these two tasks are integrated within a single agent, the LLM consistently fails to produce correct outputs, even when provided with expert-defined rules and syntactic templates. These findings highlight that, in complex domain-specific applications, effective task decomposition is the cornerstone of reliable system performance, wherein the complexity of each subtask must match the reasoning capacities of LLMs.
    
    \item The proposed multi-agent system is powered by the lightweight Llama-3.3 70B Instruct model, which enables local deployment on standard workstations and avoids reliance on high-end cloud infrastructure. Additionally, the cost analysis reveals that, despite increased token usage due to multi-agent communication, the overall running cost of the proposed system remains highly competitive relative to ChatGPT and Gemini. This balance of reliability, affordability, and deployability provides a clear path for its adoption in real-world engineering practices. 
    
    \item The proposed agentic system has the potential to substantially improve automation and efficiency in engineering workflows. For instance, generating the structural analysis code for a 3-2-3 frame requires only 269.2 seconds using the proposed system, compared to 17 minutes and 24 seconds through manual coding. Beyond time efficiency, the system supports a fully automated workflow: users provide a natural language description of the structural problem, after which the system handles the modeling and analysis process. This capability broadens the accessibility of advanced structural analysis tools, enabling their practical use by students, early-career engineers, and non-specialists.
    
\end{itemize}

\section*{Data Availability Statement}
Some or all data, models, or code that support the findings of this study are available from the corresponding author upon reasonable request.

\bibliographystyle{ascelike}  
\bibliography{references}  




\clearpage
\section*{Appendix A: A representative example of OpenSeesPy code generated by the code translation and model validation agents}
\label{appendix:translation and validation}

\begin{figure*}[htbp]
\centering
\includegraphics[width=0.86\textwidth]{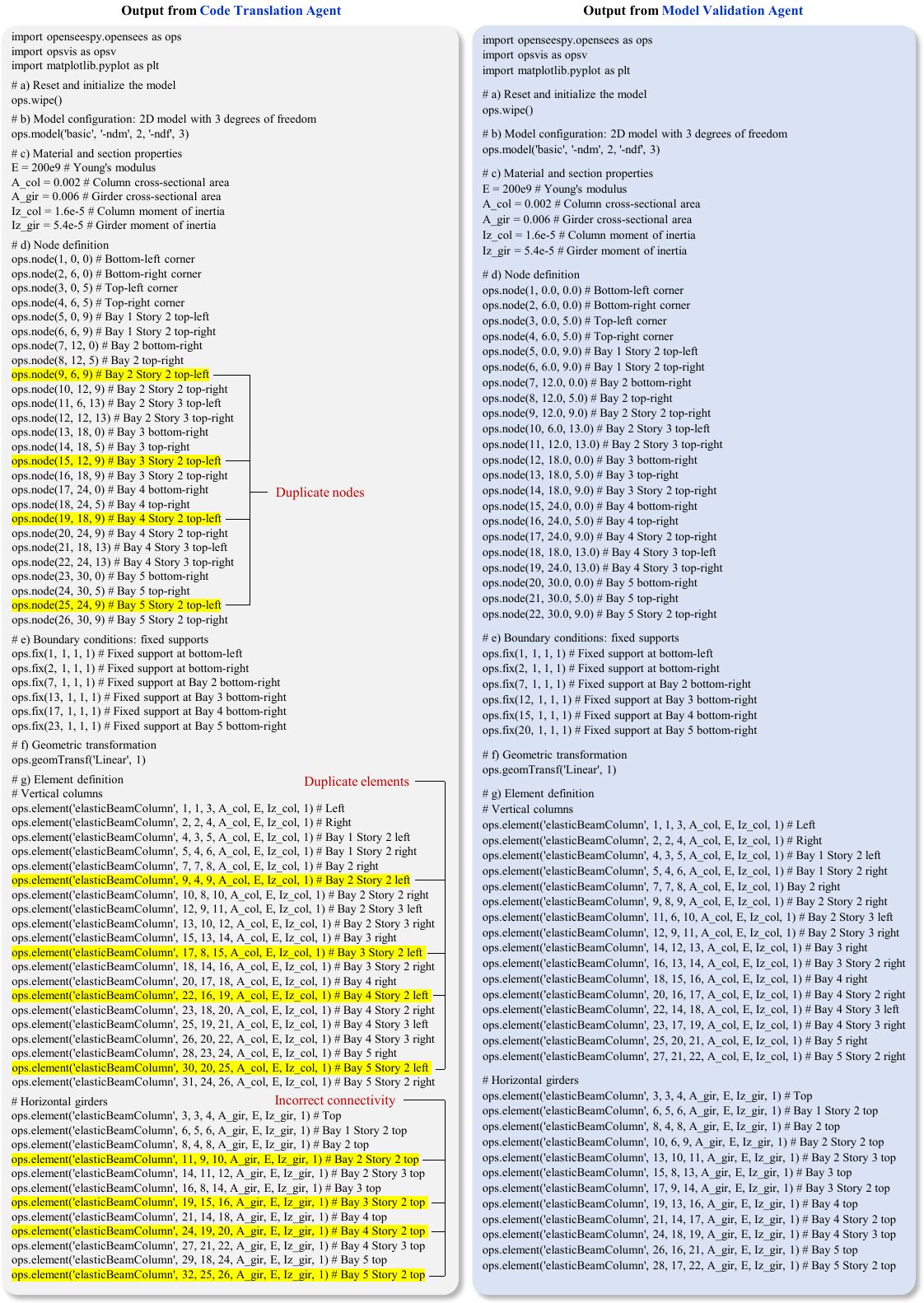}
\label{FigureS1}
\end{figure*}

\clearpage 
\section*{Appendix B: An illustrative example of the complete code generated by the multi-agent system}
\label{appendix:complete}

\begin{figure*}[htbp]
\centering
\includegraphics[width=0.87\textwidth]{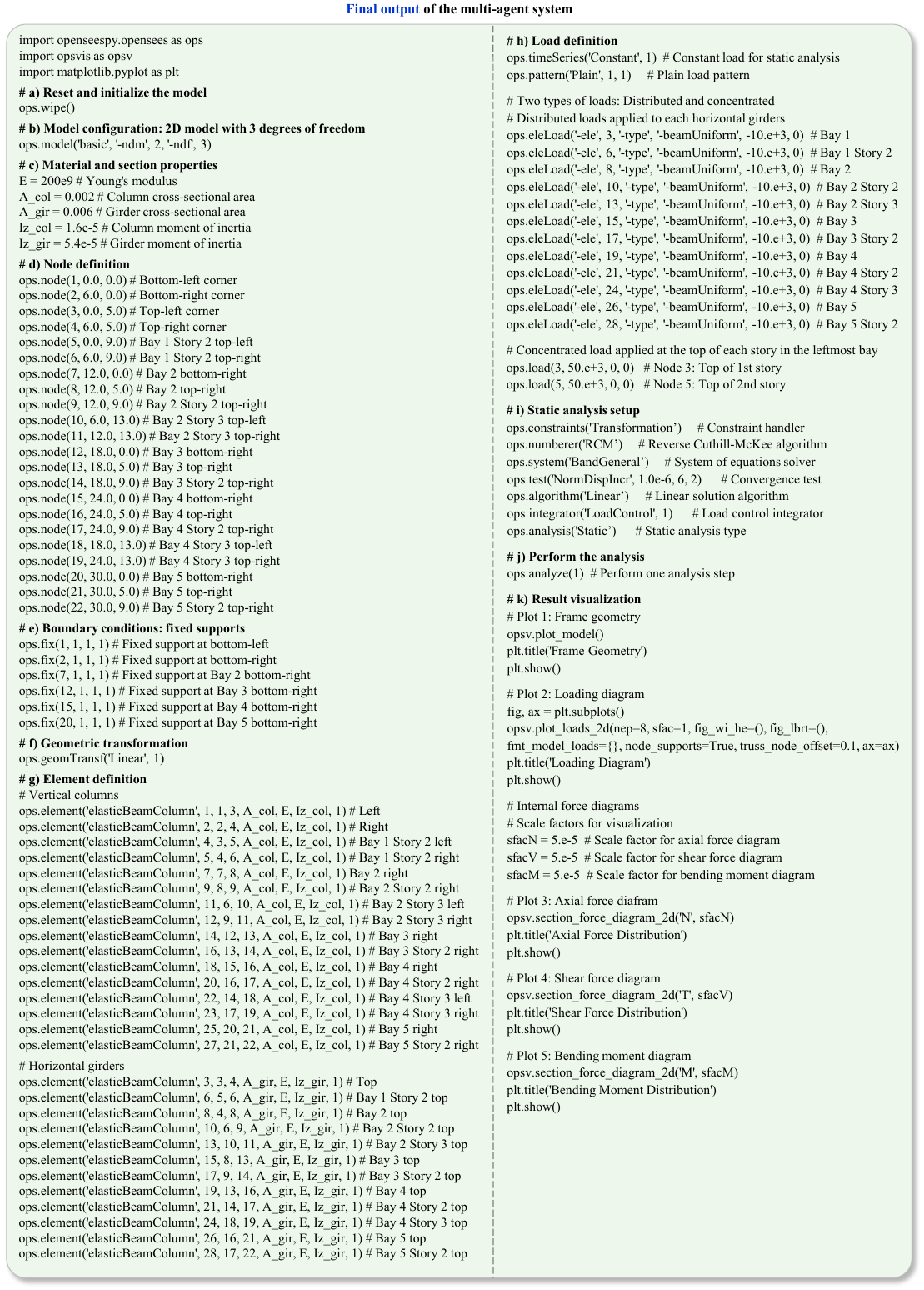}
\label{FigureS2}
\end{figure*}

\end{document}